\documentclass[10pt,twocolumn,letterpaper]{article}
\usepackage[T1]{fontenc}

\PassOptionsToPackage{table}{xcolor}
\usepackage{cvpr}

\usepackage{tabularx}
\usepackage{multirow}
\usepackage{stfloats}

\definecolor{PaperTableHeader}{HTML}{E6E6E6}
\definecolor{PaperAODTint}{HTML}{D7F6FF}
\newcommand{\ours}{AOD}
\providecommand{\Description}[1]{}

\makeatletter
\g@addto@macro\normalsize{%
  \setlength\abovedisplayskip{4pt}%
  \setlength\belowdisplayskip{4pt}%
  \setlength\abovedisplayshortskip{0pt plus 2pt}%
  \setlength\belowdisplayshortskip{4pt plus 2pt minus 2pt}%
  \setlength\jot{2pt}%
}
\makeatother
\definecolor{cvprblue}{rgb}{0.21,0.49,0.74}
\usepackage[pagebackref,breaklinks,colorlinks,allcolors=cvprblue]{hyperref}

\def\paperID{0000}
\def\confName{CVPR}
\def\confYear{2026}

\title{Ask on Demand: Targeted Peer Communication for Parallel Agentic Systems}
\author{
\textbf{Haofu Yang}$^{1,3,*}$,\quad
\textbf{Hongyang Yin}$^{2,3,*}$,\quad
\textbf{Ran Bi}$^{3,\dagger}$,\quad
\textbf{Faguo Wu}$^{1,\dagger}$\\
\textbf{Peng Xue}$^{3}$,\quad
\textbf{Feng Zhou}$^{3}$,\quad
\textbf{Han Zhang}$^{3}$\\[2pt]
$^{1}$Beihang University \quad
$^{2}$Tsinghua University \\
$^{3}$China Telecom Artificial Intelligence Technology (Beijing) Co., Ltd.
}
\hypersetup{pdfauthor={Haofu Yang, Hongyang Yin, Ran Bi, Faguo Wu, Peng Xue, Feng Zhou, Han Zhang}}

\begin{document}
\maketitle

\begingroup
\renewcommand{\thefootnote}{*}
\footnotetext[1]{Equal contribution.}
\renewcommand{\thefootnote}{\ensuremath{\dagger}}
\footnotetext[2]{Corresponding authors. Email: bir3@chinatelecom.cn} 
\endgroup

\begin{abstract}

Parallel LLM agents tackle complex tasks by exploring multiple trajectories, yet useful discoveries remain isolated in individual execution histories. Existing sharing methods often select or summarize information before other agents' needs become clear, potentially overlooking details that become important later. We propose Ask on Demand (AOD), a communication framework that enables agents to seek targeted peer assistance when specific information needs arise. Upon an approved request, AOD retrieves relevant evidence from peer execution histories, selects a suitable peer, and generates a response tailored to the request. To determine when such assistance is worthwhile, we train a lightweight gate using paired team continuations, with supervision derived from final team outcomes and inference cost.
Experiments on AppWorld and WebShop demonstrate that AOD consistently improves task performance over independent parallel execution while achieving lower inference cost
than existing information-sharing baselines.  These findings highlight the benefits of aligning peer communication with agents' evolving information needs to achieve more effective and efficient collaboration.

\end{abstract}

\section{Introduction}
\label{sec:introduction}

% Blue text marks provisional method/evaluation details and missing results.
% Remove editorial markers only after the corresponding claims are confirmed.

% P1: Why is cross-agent information reuse necessary?

As large language model (LLM) agents increasingly move beyond information retrieval to acting in interactive digital environments, they face complex, long-horizon tasks spanning website navigation, interactions with online services and APIs, cross-application workflows, and software engineering~\cite{yao2023react,shinn2023reflexion,wang2024voyager,trivedi2024appworld,yang2024sweagent}. Parallel execution offers a promising way to scale such tasks by allowing multiple agents to explore different trajectories toward the same objective~\cite{zhu2025scaling,lee2026aggagent}. Along these trajectories, agents accumulate complementary evidence, discovered constraints, successful tool-use patterns, and informative failed attempts. Yet these discoveries remain scattered across private execution histories and are not automatically available to other agents making related decisions~\cite{hu2026agentfugue}. This isolation can lead agents to repeatedly inspect the same resources or rediscover failures already encountered by their peers~\cite{fioresi2026learning}. Simply increasing communication, however, is not an ideal remedy: frequent or indiscriminate sharing incurs additional context and inference costs and may steer otherwise diverse trajectories toward the same salient, potentially misleading leads~\cite{zhang2025agentprune,hu2026agentfugue,ann2026interaction}. Thus, parallel agentic systems face a fundamental challenge: enabling cross-agent information reuse without sacrificing the benefits of independent exploration.

% P2: Why should information reuse be driven by emerging needs?

To mitigate this challenge, recent work has explored
selective information-sharing mechanisms that enable
parallel agents to reuse intermediate discoveries without
continuously exchanging their full execution histories.
For example, LTS selects intermediate steps for admission
to shared memory~\cite{fioresi2026learning}, while AgentFugue
compresses completed trajectory segments into episode notes
for later intent-driven retrieval~\cite{hu2026agentfugue}.
However, deciding what information to preserve or share
at the time of discovery requires anticipating the future
needs of other agents. Crucially, the value of an intermediate
discovery depends on the evolving information needs of its
potential recipients. A finding that appears irrelevant
early in another agent's trajectory may become critical
many steps later, when that agent encounters a specific
uncertainty or constraint. Consequently, selecting or
compressing information in advance risks overlooking
details whose relevance only becomes apparent
later~\cite{yan2025gam}. This observation suggests a
different principle: information reuse should be driven
by the recipient's current needs, rather than solely
by predictions of future utility made when information
is produced. This raises a central question:
\textbf{How can parallel agents access relevant peer
discoveries when specific information needs arise,
while avoiding unnecessary communication?}

% P3: How does AOD enable demand-driven information reuse?

\begin{figure}[t]
  \centering
  \includegraphics[width=\linewidth]{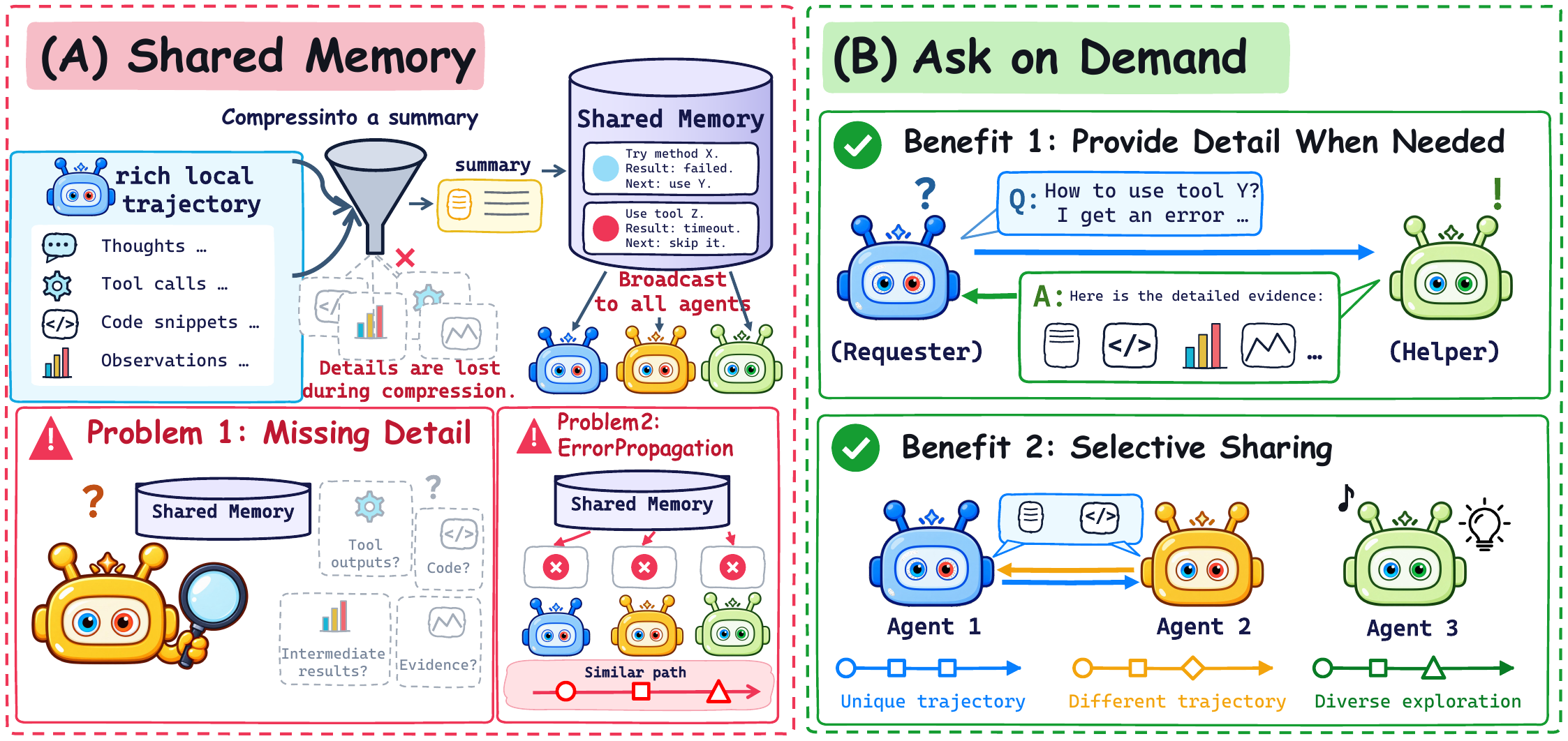}
    \caption{\textbf{Comparison of information sharing} in parallel agentic systems.
(a) Shared memory may lose useful details through advance summarization
and propagate misleading information.
(b) Ask on Demand (AOD) enables targeted access to peer evidence
when needed while preserving independent exploration.}
  \label{fig:askroute-overview}
\end{figure}

Motivated by this insight, we propose
\textbf{Ask on Demand (AOD)}, a demand-driven communication
framework for parallel agentic systems. Inspired by how
human teams seek relevant expertise when needed~\cite{faraj2000coordinating},
AOD enables agents to request assistance based on their
emerging information needs, rather than relying solely
on discoveries selected for sharing in advance.
As illustrated in
\hyperref[fig:askroute-overview]{\underline{Figure~\ref*{fig:askroute-overview}}},
upon an approved request, AOD retrieves relevant evidence
from peer execution histories to identify an appropriate
agent to consult. The selected peer then draws on its local
history to provide a response customized to the request.
This design offers two advantages. First, accessing peer
evidence at the point of need can recover details that
might be omitted by advance summarization. Second,
directing requests to relevant peers limits unnecessary
information exchange without broadcasting discoveries
to the entire team. In this way, AOD enables effective
information reuse through targeted, selective communication.

% P4: Learning when to enable help-seeking.

However, enabling on-demand communication does not by itself tell
agents when asking is worthwhile. The value of a request depends
on its effect on subsequent task execution, rather than on the
presence of an information gap alone. We therefore learn a lightweight
help-seeking policy from the estimated downstream payoff of asking.
Following rollout-based communication learning~\cite{wang2026learning},
we compare paired team continuations from the same execution
state, with the same proposed request approved or denied.
A fixed rule selects one agent's result as the team output
using self-assessed confidence. The paired comparisons
yield binary training labels that prioritize the selected
result's task score and favor lower remaining inference
cost when scores are tied. We use this supervision to fine-tune
the policy to identify requests worth approving.
At inference, the policy uses only the proposed request and
requester-local context to decide whether to approve it,
while the frozen task model determines what to ask and how
to use the reply.

% P5: Experimental evaluation.

We evaluate AOD on two long-horizon, interactive benchmarks:
AppWorld and WebShop, covering cross-application workflows
and interactive product search~\cite{trivedi2024appworld,yao2022webshop}.
Both benchmarks require agents to accumulate observations
and use intermediate findings to guide subsequent decisions,
providing complementary settings for studying information
reuse across parallel trajectories.
Experiments demonstrate that AOD improves task performance
while achieving a favorable balance between effectiveness
and inference cost compared with existing baselines.
Further analyses highlight the benefits of learned help-seeking
decisions and relevant peer assistance, and examine exploration
behavior, gate supervision, and scalability to larger teams.

Our main contributions are summarized as follows:
\begin{itemize}
    \item \textbf{Demand-Driven Peer Communication.}
    We propose Ask on Demand (AOD), a communication framework that enables parallel agents to seek targeted assistance based on their evolving information needs, rather than relying on information shared in advance.

    \item \textbf{Outcome-Guided Help Seeking.}
    We develop a lightweight gate that learns when peer assistance is worthwhile by comparing the effects of approving and denying requests on final team outcomes.

    \item \textbf{Experimental Validation.}
    Experiments on AppWorld and WebShop demonstrate the effectiveness of AOD in improving task performance while balancing inference cost. Further analyses validate its key components and examine scalability to larger teams.
\end{itemize}

% Label the panels as schematic unless grounded in actual execution logs.
% Do not depict summary misses or trajectory convergence as observed baseline
% failures without trace evidence.

\section{Related Work}
\label{sec:related-work}

\paragraph{Parallel Reasoning in Agentic Systems.}
Parallel reasoning improves exploration during inference by generating multiple candidate trajectories. Self-consistency and repeated sampling demonstrate the benefits of aggregating and verifying candidate solutions~\cite{wang2023selfconsistency,brown2024monkeys}. This approach extends to agents that interact with tools and environments, where completed trajectories can be selected or combined to produce a final solution~\cite{zhu2025scaling,lee2026aggagent}. Recent studies further enable agents to reuse discoveries during execution. LTS selectively stores intermediate results in shared memory for later retrieval~\cite{fioresi2026learning}, while AgentFugue enables agents to access relevant peer episodes through a shared reasoning hub~\cite{hu2026agentfugue}. ExComm combines targeted belief correction with explicit diversification of exploration plans~\cite{song2026excomm}. Building on these studies, our work focuses on accessing relevant discoveries from peer trajectories according to agents' emerging information needs.

\paragraph{Communication in Multiagent Systems.}
Earlier multi-agent reinforcement learning methods learn when to
communicate through gating~\cite{singh2019learning} and whom to address
through targeted attention~\cite{das2019tarmac}.
LLM agent frameworks support collaboration through role assignments, message exchange, and structured workflows~\cite{wu2024autogen,li2023camel,hong2024metagpt,qian2024chatdev}. Multiagent debate further enables models to revise their answers through repeated exchanges of responses and reasoning~\cite{du2024debate}. To improve communication efficiency, existing methods optimize interactions through pruning and dynamic routing. AgentPrune removes unnecessary communication links~\cite{zhang2025agentprune}, while DyTopo dynamically matches agents' needs with available assistance~\cite{lu2026dytopo}. At the workflow level, GPTSwarm optimizes prompts and graph connectivity~\cite{zhuge2024gptswarm}, while DAAO adapts agentic workflows to predicted query difficulty~\cite{su2026difficulty}. Other studies investigate learned clarification at workflow transitions~\cite{lin2026agentask} and communication guided by agents' beliefs about their teammates and potential action conflicts~\cite{wang2026cobelworld}. In contrast, \ours{} focuses on agents exploring the same task in parallel, enabling them to request relevant evidence from peer execution histories when needed. This allows agents to reuse complementary discoveries while limiting communication overhead and preserving independent exploration.

\paragraph{Adaptive Information Retrieval.}
Adaptive retrieval methods condition information access on the progress of generation or task execution. FLARE retrieves when anticipated content contains low-confidence tokens~\cite{jiang2023active}, while Self-RAG learns to retrieve and assess supporting passages~\cite{asai2024selfrag}. ITER further incorporates interaction history and pre-search reasoning into retrieval~\cite{chen2026iter}. AOD studies information access across parallel agents, using concrete requests to retrieve peer evidence and team outcomes to learn whether an exchange is worthwhile.

\section{Method}
\label{sec:method}

\begin{figure*}[t]
  \centering
  \includegraphics[width=\linewidth]{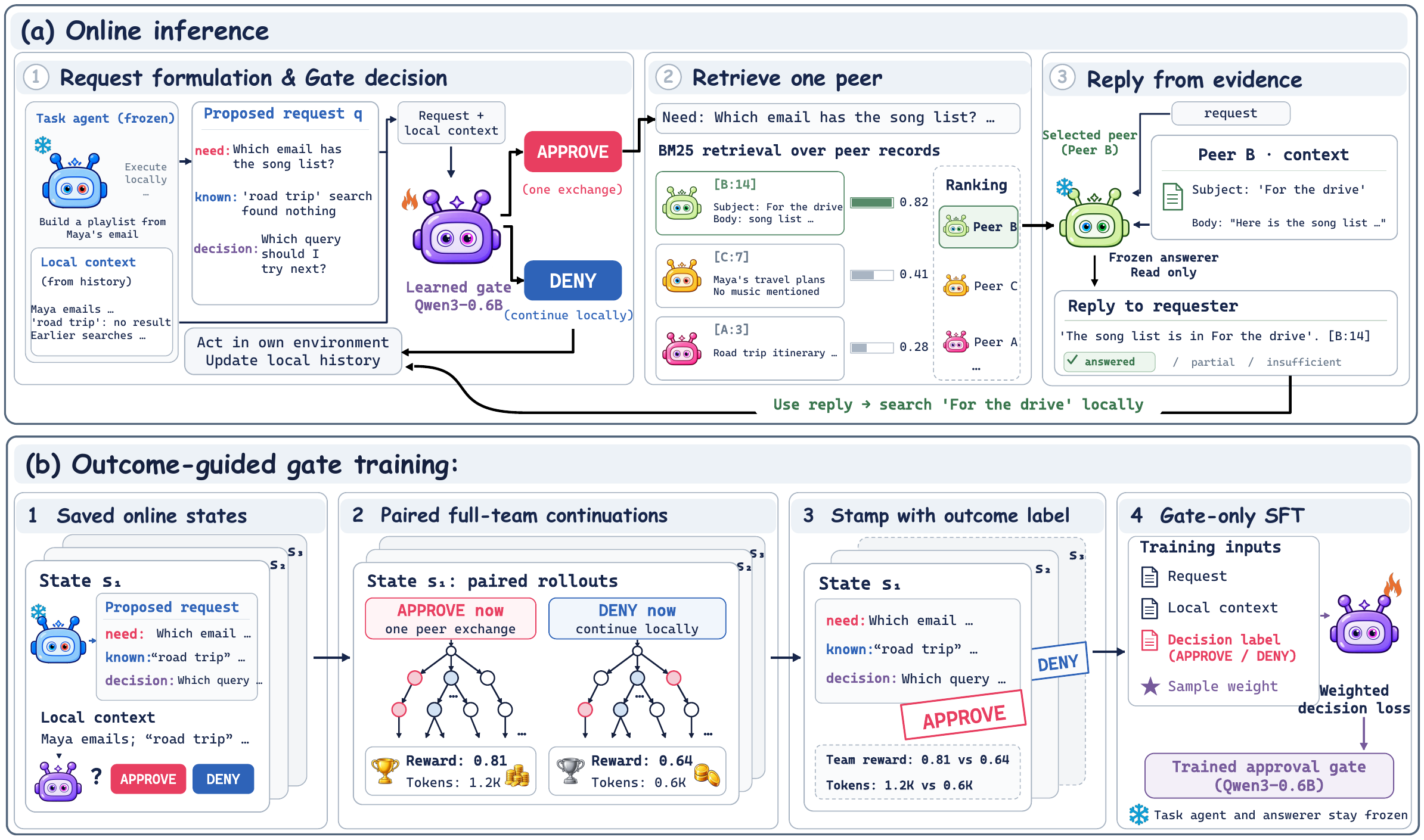}

\caption{\textbf{Overview of Ask on Demand (AOD).}
(a) \textbf{Online inference.} A learned gate approves requests,
after which AOD selects a relevant peer to provide an
evidence-grounded response.
(b) \textbf{Gate training.} Paired team continuations compare
approving and denying the same request, using final team
outcomes and inference cost to supervise the gate.
Rewards and token costs are illustrative means over paired
continuations, with costs measured after branching.}

  \label{fig:method}
\end{figure*}

We propose \emph{Ask on Demand} (AOD), a framework that enables
agents to request peer discoveries relevant to their current
decisions (Figure~\ref{fig:method}). When an agent needs information, it formulates a structured
request that specifies what is missing and which decision
the answer could inform.
Given that a request incurs inference cost and its downstream
value is unknown in advance, we introduce a lightweight 0.6B-parameter gate to decide whether
to approve it before any peer is consulted. Upon approval, AOD uses this request to retrieve relevant
completed records, select a single peer, and generate a targeted
answer that cites that peer's observations.

To learn which requests are worth approving, we restore the same
team state and compare full-team continuations that approve or
deny the same request. These comparisons yield binary labels
that prioritize final team performance and favor lower inference
cost when performance is tied. The gate thus learns from
downstream team outcomes, yet at inference it sees only the
request and the requester's local context. All task-solving and
peer-answering models remain frozen; only the gate is trained.

\subsection{Problem Formulation}
\label{sec:formulation}

\noindent\paragraph{\textbf{Team Setting.}}
We consider a team of $K$ agents solving the same task $u$
in separate environment instances. Each agent $k$ maintains its
own interaction history $h_{k,t}$, including any peer replies
received so far.

\noindent\paragraph{\textbf{Communication Decision.}}
When agent $i$ proposes a peer request $q_{i,t}$
(Section~\ref{sec:ask}), the help-seeking gate evaluates it
before execution. The gate thus decides whether this specific request is worth executing. Its input
$x_{i,t}$ contains the proposed request and the requester's local context. The gate is shared across agents and defines a binary policy:
\begin{equation}
    \pi_\theta(a\mid x_{i,t}),
    \qquad
    a\in\{\mathrm{APPROVE},\mathrm{DENY}\}.
\end{equation}
Here, $\mathrm{APPROVE}$ executes the proposed request, whereas
$\mathrm{DENY}$ rejects it and allows local execution to continue.
Denying the current request does not preclude later requests.

\noindent\paragraph{\textbf{Team Outcome.}}
At team termination, each agent $k$ has produced a trajectory
$\tau_k$ and a self-assessed confidence $c_k$~\cite{tian2023just}. A fixed selection
rule $\sigma$ selects one trajectory based on these confidence
scores. We denote the selected trajectory's normalized task
score by $R$:
\begin{equation}
    \hat{\imath}
    =\sigma\big(\{(\tau_k,c_k)\}_{k=1}^{K}\big),
    \qquad
    R=\mathcal{R}(\tau_{\hat{\imath}})\in[0,1].
\end{equation}
Benchmark-specific selection and failure rules are detailed
in the supplementary material.
We measure total inference cost as
\begin{equation}
    C=\sum_{v\in\mathcal{V}}
    \left(N_v^{\mathrm{in}}+N_v^{\mathrm{out}}\right),
\end{equation}
where $\mathcal{V}$ indexes all task and auxiliary model calls, and $N_v^{\mathrm{in}}$ and $N_v^{\mathrm{out}}$ are the input and output token counts of call $v$.

\subsection{Ask-on-Demand Communication}
\label{sec:ask}

An AOD exchange is organized around the requester's pending
decision. The request guides both peer selection and the answer
generated from the selected peer's records.

\noindent\paragraph{\textbf{Request Formulation.}}
Targeted questions can resolve information gaps that would
otherwise require an agent to guess~\cite{lin2026agentask}. AOD makes \texttt{ask\_peer} available alongside the
agent's environment tools. When the frozen task model
$\mathcal{M}$ selects it as the next action, the same
model call generates the request arguments from the
requester's interaction history $h_{i,t}$:
\begin{equation}
    q_{i,t}
    =(\mathrm{need},\mathrm{known},\mathrm{decision})
    =\mathcal{M}(h_{i,t}).
\end{equation}
Here, $\mathrm{need}$ names the missing fact or lead,
$\mathrm{known}$ summarizes the requester's relevant observations
and previously received replies while distinguishing them from
assumptions, and $\mathrm{decision}$ specifies what the requester
plans to do depending on the answer. The gate never writes or
edits $q_{i,t}$; it only decides whether the proposed request is
executed. An approved request is executed once, as written.

\noindent\paragraph{\textbf{Peer Selection.}}
Once the gate approves $q_{i,t}$, the request is matched against
the other agents' completed records. Let $P_{j,t}$ denote agent $j$'s completed records at the moment of the request. Actions still running and anything recorded later are not visible. Unlike the full history $h_{j,t}$, these records do not include replies received from other agents. Among peers with retrieved candidates, we select the one
with the highest-scoring passage and consult only that peer:
\begin{equation}
    j^{\star}
    =\arg\max_{j\neq i}\ \max_{c\in\mathcal{C}(P_{j,t})}
    \operatorname{rel}(q_{i,t},c),
\end{equation}
where $\mathcal{C}(P_{j,t})$ contains text passages from
agent $j$'s completed records, and $\operatorname{rel}$ is a BM25-based
relevance score~\cite{robertson2009probabilistic}. If no positive-scoring candidates are found, AOD returns
an insufficient reply without consulting a peer.

\noindent\paragraph{\textbf{Evidence-Grounded Answering.}}
Let $E_{j^\star,t}$ denote the evidence excerpts retrieved
from $P_{j^\star,t}$ within the evidence token budget.
As in retrieval-augmented generation~\cite{lewis2020retrieval},
the selected peer's frozen answering model
$\mathcal{M}_{\mathrm{ans}}$ generates a concise reply $r_{i,t}$ conditioned on these excerpts:
\begin{equation}
    r_{i,t}
    =\mathcal{M}_{\mathrm{ans}}
      \big(u,q_{i,t},E_{j^\star,t}\big).
\end{equation}
The stated need
and decision guide which findings to explain and how they
can help the requester proceed. The reply includes citations and exact excerpts, so the
requester can inspect the evidence behind the
answer~\cite{gao2023enabling}. Answering the request does not trigger new environment actions
or interrupt the peer's ongoing execution. If the available
records are insufficient, the peer can give a partial answer or indicate that it cannot answer. The requester remains
responsible for verifying suggestions and acting in its own
environment.

An answered request need not be worthwhile: its value depends
on the resulting changes in team performance and inference
cost~\cite{becker2009analyzing}. We estimate these effects next.

\subsection{Estimating the Value of a Request}
Building on rollout-based communication learning~\cite{wang2026learning},
we compare future
team reward and remaining inference cost across paired full-team
continuations from the same state. The proposed request is held
fixed: one continuation approves it, and the other denies it.
Both continuations handle later requests under the same policy,
so the comparison measures the marginal value of executing this
request now.
\noindent\paragraph{\textbf{Request Collection and Replay.}}
We first run each task in the training set to completion
and record the full team execution. These runs follow a fixed
policy $\pi_b$: task models proceed with the task and propose
requests naturally, while the original Qwen3-0.6B gate~\cite{qwen2025qwen3} approves
or denies them. We keep a limited number of requests per task,
chosen without using their continuation outcomes. 
For each kept request $q=q_{i,t}$, let $s$ denote the full team state
immediately before the gate evaluates it: all agents' histories,
environment states, completed records, and remaining budgets.
We restore $s$ by replaying the recorded team prefix, reusing
previous model outputs rather than regenerating them, and verify
that both the state and the pending request match the recording
before branching.
\noindent\paragraph{\textbf{Paired Team Continuations.}}
From $s$, we create two branches that differ only in the decision
on $q$. The APPROVE branch executes $q$ through the communication
mechanism of Section~\ref{sec:ask}; the DENY branch rejects it,
and the requester continues locally. For later requests, both branches
use the same rollout policy $\pi_p$, which approves each subsequent
request with probability $p$ (default $0.5$), until team termination
or budget exhaustion. This probability affects later requests, not
the paired APPROVE/DENY intervention on the initial request.
This allows the current request to be evaluated across different
future communication decisions while retaining later opportunities
for peer help in both branches. We record
the final team reward $R$ and the inference cost $C_{\mathrm{suf}}$
incurred after $s$.
\noindent\paragraph{\textbf{Value Estimation.}}
Under the continuation policy $\pi_p$, we define the value
of executing $q$ as its expected effect on team reward and
remaining inference cost:
\begin{equation}
    \begin{aligned}
    \Delta R_p(s,i,q)
    &= \mathbb{E}_{\pi_p}[R\mid s,i,q,\mathrm{APPROVE}] \\
    &\quad - \mathbb{E}_{\pi_p}[R\mid s,i,q,\mathrm{DENY}],
    \end{aligned}
\end{equation}
\begin{equation}
    \begin{aligned}
    \Delta C_p(s,i,q)
    &= \mathbb{E}_{\pi_p}
      [C_{\mathrm{suf}}\mid s,i,q,\mathrm{APPROVE}] \\
    &\quad - \mathbb{E}_{\pi_p}
      [C_{\mathrm{suf}}\mid s,i,q,\mathrm{DENY}].
    \end{aligned}
\end{equation}
We estimate both quantities using an adaptive number of valid
paired continuations. For each request, we start with five pairs
and add five whenever the estimates remain unstable, up to
twenty pairs (see the supplementary material). The two branches of
each pair share a random seed~\cite{glasserman1992common}.
Let $n\in\{5,10,15,20\}$ denote the final number of valid pairs.
Averaging all $n$ paired differences gives
\begin{equation}
    \widehat{\Delta R}_p(s,i,q)
    = \frac{1}{n}\sum_{\ell=1}^{n}
    \big(
        R_{\mathrm{APPROVE}}^{(\ell)}
        - R_{\mathrm{DENY}}^{(\ell)}
    \big),
\end{equation}
\begin{equation}
    \widehat{\Delta C}_p(s,i,q)
    = \frac{1}{n}\sum_{\ell=1}^{n}
    \big(
        C_{\mathrm{suf},\mathrm{APPROVE}}^{(\ell)}
        - C_{\mathrm{suf},\mathrm{DENY}}^{(\ell)}
    \big).
\end{equation}

These estimates provide supervision under $\pi_p$;
sensitivity to this continuation policy is examined in
Section~\ref{sec:sensitivity}.

\subsection{Learning the Help-Seeking Gate}
\label{sec:gate}

We derive binary approval labels from the paired estimates
of team reward and remaining inference cost. We then train
the gate to predict these labels from the proposed request
and the requester's local context.

\noindent\paragraph{\textbf{Supervision from Outcomes.}}
Each collected request receives one binary label. Team reward
determines the label; remaining cost matters only when the
reward is unchanged. For the fixed rollout probability $p$, write
$\widehat{\Delta R}=\widehat{\Delta R}_p(s,i,q)$ and
$\widehat{\Delta C}=\widehat{\Delta C}_p(s,i,q)$. We assign:
\begin{equation}
y(s,i,q)=
\begin{cases}
\mathrm{APPROVE},
  & \widehat{\Delta R}>\varepsilon,\\[2pt]
\mathrm{APPROVE},
  & |\widehat{\Delta R}|\leq\varepsilon
    \ \land\ \widehat{\Delta C}<-\varepsilon,\\[2pt]
\mathrm{DENY},
  & \text{otherwise},
\end{cases}
\end{equation}
where $\varepsilon=10^{-9}$ is a numerical tolerance
for floating-point comparisons.
A request that raises team reward is labeled
$\mathrm{APPROVE}$ even if it costs more; when the reward
is tied, it is labeled $\mathrm{APPROVE}$ only if it lowers
the remaining cost.

\noindent\paragraph{\textbf{Training Set.}}
Each training example $(x_{i,t},y,w)$ pairs the gate input
$x_{i,t}$ with its approval label $y$ and a sample weight $w$.
Here, $x_{i,t}$ contains the proposed request $q_{i,t}$ and the
requester-visible information available when state $s$ was
recorded. We weight each example according to the number of
valid pairs and the variability of their outcomes. The base
weight is
\begin{equation}
\label{eq:base-weight}
    \widetilde w
    =\operatorname{clip}_{[0.1,1]}
    \left(
    \frac{n/(n+3)}
    {1+\mathrm{SE}_{R}/0.1+\mathrm{SE}_{\mathrm{relC}}/0.25}
    \right).
\end{equation}
Requests that remain unstable after twenty pairs receive an
additional reduction:
\begin{equation}
\label{eq:sample-weight}
w=
\begin{cases}
    0.25\,\widetilde w,
    & n=20\ \text{and still unstable},\\[2pt]
    \widetilde w,
    & \text{otherwise}.
\end{cases}
\end{equation}
The standard-error terms and stability criteria are defined
in the supplementary material. Each request retains a binary
label determined by its mean reward and cost differences.
We retain all $\mathrm{APPROVE}$ examples and subsample
$\mathrm{DENY}$ examples to form the training set $\mathcal{D}$.

\noindent\paragraph{\textbf{Learning the Gate.}}
We fully fine-tune Qwen3-0.6B to predict
$\mathrm{APPROVE}$ or $\mathrm{DENY}$ for each proposed
request. Let $z_m$ denote the output token sequence encoding
label $y_m$, with length $L_m$, and let $p_\theta$ denote
the model's next-token distribution over the full vocabulary.
We minimize the weighted cross-entropy loss:
\begin{equation}
\mathcal{L}(\theta)
= -\frac{1}{\sum_{m\in\mathcal{D}}w_m}
  \sum_{m\in\mathcal{D}} w_m
  \sum_{r=1}^{L_m}
  \log p_{\theta}(z_{m,r}\mid x_m,z_{m,<r}).
\end{equation}
Here, $z_{m,<r}$ denotes the output tokens preceding position $r$.
Only the tokens in the decision output contribute to the loss.
All task-solving and peer-answering models remain frozen;
only the gate is updated.

\noindent\paragraph{\textbf{Inference.}}
At inference, the trained gate evaluates each proposed
request using $x_{i,t}$. An $\mathrm{APPROVE}$ decision
executes the request through the communication mechanism
in Section~\ref{sec:ask}; a $\mathrm{DENY}$ decision rejects
it and allows local execution to continue. Gate decisions require neither
paired continuations nor environment rewards at inference.

% Result formatting shared with the generated tables.
\providecommand{\expval}[1]{\if\relax\detokenize{#1}\relax\texttt{xx.x}\else#1\fi}
\providecommand{\expstd}[2]{\ensuremath{#1_{\textcolor{black!55}{\scriptscriptstyle #2}}}}
\providecommand{\expna}{\textnormal{N/A}}
% BEGIN GENERATED: tables/result_numbers.tex
% Generated by experiments/derive_results.py; edit result files instead.
\providecommand{\AppScoreDelta}{\expval{+8.9}}
\providecommand{\AppCostRatio}{\expval{1.09}}
\providecommand{\WebScoreDelta}{\expval{+2.8}}
\providecommand{\WebCostRatio}{\expval{0.92}}
\providecommand{\WebSRDelta}{\expval{+3.0}}
\providecommand{\GateWebAskDelta}{\expval{+0.50}}
\providecommand{\GateWebCostChange}{\expval{-8.03}}
\providecommand{\GateWebSRDelta}{\expval{+0.5}}
% END GENERATED: tables/result_numbers.tex

% Keep the experimental paragraphs top-aligned when a column is short.
\begingroup
\raggedbottom
\setlength{\emergencystretch}{2em}
\section{Experiments}
\label{sec:experiments}

We evaluate \ours{} on two interactive benchmarks, focusing on task performance, inference efficiency, and exploration diversity in parallel agentic systems. Our experiments examine the effectiveness of on-demand communication and the contributions of its key components, including the learned help-seeking gate, peer selection, and evidence-grounded answering.

\subsection{Experimental Setup}
\label{sec:experimental_setup}

\noindent\paragraph{\textbf{Benchmarks}.}
We evaluate \ours{} on two interactive benchmarks.
\textbf{AppWorld}~\cite{trivedi2024appworld} evaluates agents on cross-application tasks involving API interactions in stateful environments. We use a subset of 120 tasks spanning 40 scenarios.
\textbf{WebShop}~\cite{yao2022webshop} evaluates agents on product search and purchasing tasks based on natural-language requirements. We evaluate on 200 shopping tasks.
Further details on task selection and data splits are provided in the supplementary material.

% BEGIN GENERATED: tables/main_results.tex
% Generated by experiments/render_tables.py; edit the CSV, not this file.
\begin{table*}[!t]
\centering
\caption{\textbf{Task performance and oracle outcomes}
on two interactive benchmarks. Improvement is AOD minus the
strongest baseline in each column for the same task model,
in points or percentage points. Entries show means, with standard
deviations over three seeds in subscripts.}
\label{tab:main_results}
\footnotesize
\renewcommand{\arraystretch}{1.18}
\setlength{\tabcolsep}{1.1pt}
\begin{tabular}{p{0.095\textwidth}p{0.145\textwidth}*{8}{>{\centering\arraybackslash}p{\dimexpr(.76\textwidth-20\tabcolsep)/8\relax}}}
\toprule[0.7pt]
\rowcolor{PaperTableHeader}
&
& \multicolumn{4}{c}{\textbf{AppWorld}}
& \multicolumn{4}{c}{\textbf{WebShop}} \\
\rowcolor{PaperTableHeader}
& & &
& \multicolumn{2}{c}{\textbf{Oracle@3}}
& &
& \multicolumn{2}{c}{\textbf{Oracle@3}} \\
\rowcolor{PaperTableHeader}
\multirow{-3}{*}{\textbf{Model}} & \multirow{-3}{*}{\textbf{Method}}
& \multirow{-2}{*}{\textbf{TGC}$\uparrow$}
& \multirow{-2}{*}{\textbf{SGC}$\uparrow$}
& \textbf{TGC}$\uparrow$ & \textbf{SGC}$\uparrow$
& \multirow{-2}{*}{\textbf{TS}$\uparrow$}
& \multirow{-2}{*}{\textbf{SR}$\uparrow$}
& \textbf{TS}$\uparrow$ & \textbf{SR}$\uparrow$ \\
\midrule[0.35pt]
\multirow{10}{0.095\textwidth}{\centering\textbf{\shortstack{Qwen3.5-\\35B-A3B}}}
& \multicolumn{9}{c}{\cellcolor{PaperTableHeader}\textbf{Single agent}} \\
& Base & \expval{\expstd{43.33}{2.49}} & \expval{\expstd{15.00}{0.00}} & - & - & \expval{\expstd{62.03}{2.75}} & \expval{\expstd{40.17}{2.54}} & - & - \\
& ReAct & \expval{\expstd{49.44}{3.31}} & \expval{\expstd{27.50}{0.00}} & - & - & \expval{\expstd{66.75}{3.17}} & \expval{\expstd{41.17}{2.82}} & - & - \\
& \multicolumn{9}{c}{\cellcolor{PaperTableHeader}\textbf{Parallel agents}} \\
& Independent & \expval{\expstd{53.89}{5.36}} & \expval{\expstd{30.00}{8.49}} & \expval{\expstd{67.22}{3.57}} & \expval{\expstd{44.17}{4.32}} & \expval{\expstd{67.45}{0.96}} & \expval{\expstd{42.50}{1.39}} & \expval{\expstd{72.41}{0.77}} & \expval{\expstd{49.67}{0.96}} \\
& PeriodicShare & \expval{\expstd{54.72}{8.24}} & \expval{\expstd{30.00}{8.17}} & \expval{\expstd{68.33}{3.77}} & \expval{\expstd{48.33}{7.12}} & \expval{\expstd{68.38}{0.09}} & \expval{\expstd{42.67}{0.53}} & \expval{\expstd{72.17}{0.65}} & \expval{\expstd{47.67}{0.53}} \\
& LTS (Adapted) & \expval{\expstd{54.44}{5.19}} & \expval{\expstd{30.83}{4.32}} & \expval{\expstd{69.72}{1.09}} & \expval{\expstd{\underline{50.00}}{2.83}} & \expval{\expstd{\underline{69.16}}{2.05}} & \expval{\expstd{\underline{43.50}}{3.49}} & \expval{\expstd{\underline{73.68}}{0.89}} & \expval{\expstd{\underline{50.00}}{1.60}} \\
& Agentfugue (Adapted) & \expval{\expstd{\underline{59.72}}{2.93}} & \expval{\expstd{\underline{32.50}}{4.33}} & \expval{\expstd{\underline{71.67}}{3.00}} & \expval{\expstd{46.67}{8.78}} & \expval{\expstd{67.80}{0.73}} & \expval{\expstd{43.17}{2.18}} & \expval{\expstd{71.78}{0.13}} & \expval{\expstd{49.50}{0.80}} \\
& \cellcolor{PaperAODTint}\textbf{AOD} & \cellcolor{PaperAODTint}\expval{\expstd{\mathbf{62.78}}{2.68}} & \cellcolor{PaperAODTint}\expval{\expstd{\mathbf{38.33}}{2.89}} & \cellcolor{PaperAODTint}\expval{\expstd{\mathbf{73.61}}{0.48}} & \cellcolor{PaperAODTint}\expval{\expstd{\mathbf{53.33}}{1.44}} & \cellcolor{PaperAODTint}\expval{\expstd{\mathbf{70.27}}{1.06}} & \cellcolor{PaperAODTint}\expval{\expstd{\mathbf{45.50}}{1.73}} & \cellcolor{PaperAODTint}\expval{\expstd{\mathbf{74.92}}{1.45}} & \cellcolor{PaperAODTint}\expval{\expstd{\mathbf{51.33}}{1.26}} \\
\cmidrule(l){2-10}
& \cellcolor{PaperTableHeader}\textit{Improvement} & \cellcolor{PaperTableHeader}\expval{+3.06} & \cellcolor{PaperTableHeader}\expval{+5.83} & \cellcolor{PaperTableHeader}\expval{+1.94} & \cellcolor{PaperTableHeader}\expval{+3.33} & \cellcolor{PaperTableHeader}\expval{+1.11} & \cellcolor{PaperTableHeader}\expval{+2.00} & \cellcolor{PaperTableHeader}\expval{+1.24} & \cellcolor{PaperTableHeader}\expval{+1.33} \\
\midrule[0.35pt]
\multirow{10}{0.095\textwidth}{\centering\textbf{\shortstack{Qwen3.5-\\4B}}}
& \multicolumn{9}{c}{\cellcolor{PaperTableHeader}\textbf{Single agent}} \\
& Base & \expval{\expstd{16.94}{4.28}} & \expval{\expstd{2.50}{2.50}} & - & - & \expval{\expstd{31.32}{1.56}} & \expval{\expstd{15.33}{0.76}} & - & - \\
& ReAct & \expval{\expstd{13.89}{2.55}} & \expval{\expstd{1.67}{2.89}} & - & - & \expval{\expstd{42.84}{3.39}} & \expval{\expstd{24.00}{3.97}} & - & - \\
& \multicolumn{9}{c}{\cellcolor{PaperTableHeader}\textbf{Parallel agents}} \\
& Independent & \expval{\expstd{27.50}{2.50}} & \expval{\expstd{6.67}{2.89}} & \expval{\expstd{30.56}{1.73}} & \expval{\expstd{9.17}{3.82}} & \expval{\expstd{55.51}{1.95}} & \expval{\expstd{27.17}{2.75}} & \expval{\expstd{58.77}{2.72}} & \expval{\expstd{31.83}{2.84}} \\
& PeriodicShare & \expval{\expstd{24.72}{2.68}} & \expval{\expstd{7.50}{6.61}} & \expval{\expstd{28.06}{2.93}} & \expval{\expstd{9.17}{8.04}} & \expval{\expstd{\underline{61.63}}{1.37}} & \expval{\expstd{29.67}{1.61}} & \expval{\expstd{\underline{66.28}}{1.61}} & \expval{\expstd{36.33}{2.84}} \\
& LTS (Adapted) & \expval{\expstd{\mathbf{31.39}}{1.27}} & \expval{\expstd{\mathbf{12.50}}{4.33}} & \expval{\expstd{\underline{35.28}}{1.73}} & \expval{\expstd{\underline{16.67}}{1.44}} & \expval{\expstd{56.58}{0.18}} & \expval{\expstd{28.33}{2.57}} & \expval{\expstd{60.31}{0.61}} & \expval{\expstd{32.17}{2.52}} \\
& Agentfugue (Adapted) & \expval{\expstd{29.17}{0.83}} & \expval{\expstd{9.17}{2.89}} & \expval{\expstd{31.94}{1.27}} & \expval{\expstd{13.33}{1.44}} & \expval{\expstd{59.84}{2.11}} & \expval{\expstd{\underline{31.83}}{2.75}} & \expval{\expstd{64.20}{1.49}} & \expval{\expstd{\underline{37.67}}{0.76}} \\
& \cellcolor{PaperAODTint}\textbf{AOD} & \cellcolor{PaperAODTint}\expval{\expstd{\underline{31.11}}{1.73}} & \cellcolor{PaperAODTint}\expval{\expstd{\underline{10.83}}{3.82}} & \cellcolor{PaperAODTint}\expval{\expstd{\mathbf{38.06}}{1.27}} & \cellcolor{PaperAODTint}\expval{\expstd{\mathbf{18.33}}{1.44}} & \cellcolor{PaperAODTint}\expval{\expstd{\mathbf{62.90}}{0.79}} & \cellcolor{PaperAODTint}\expval{\expstd{\mathbf{32.33}}{2.02}} & \cellcolor{PaperAODTint}\expval{\expstd{\mathbf{68.71}}{0.58}} & \cellcolor{PaperAODTint}\expval{\expstd{\mathbf{39.00}}{2.50}} \\
\cmidrule(l){2-10}
& \cellcolor{PaperTableHeader}\textit{Improvement} & \cellcolor{PaperTableHeader}\expval{-0.28} & \cellcolor{PaperTableHeader}\expval{-1.67} & \cellcolor{PaperTableHeader}\expval{+2.78} & \cellcolor{PaperTableHeader}\expval{+1.66} & \cellcolor{PaperTableHeader}\expval{+1.27} & \cellcolor{PaperTableHeader}\expval{+0.50} & \cellcolor{PaperTableHeader}\expval{+2.42} & \cellcolor{PaperTableHeader}\expval{+1.33} \\
\bottomrule[0.7pt]
\end{tabular}
\end{table*}
% END GENERATED: tables/main_results.tex

\noindent\paragraph{\textbf{Baselines}.}
We compare \ours{} with three categories of baselines.
1)\underline{ {Single-agent methods}}, including
\textbf{Base} and \textbf{ReAct}~\cite{yao2023react},
which use standard agent execution and interleaved reasoning
and acting, respectively.
2) \underline{{Independent parallel execution}}, represented by
\textbf{Independent}, which runs three agents without communication.
3) \underline{{Communication-based methods}}, including
\textbf{PeriodicShare}, \textbf{LTS (Adapted)}~\cite{fioresi2026learning},
and \textbf{Agentfugue (Adapted)}~\cite{hu2026agentfugue},
which enable information exchange through periodic sharing,
shared memory, and intent-based peer retrieval, respectively.
Both adaptations use a frozen DeepSeek-V4-Pro
controller~\cite{deepseekai2026deepseekv4} without additional training;
The supplementary material describes their implementation.

\noindent\paragraph{\textbf{Evaluation Metrics}.}
We evaluate task performance and inference cost.
For task performance, we report \underline{Task Goal Completion} (TGC)
and \underline{Scenario Goal Completion} (SGC) on AppWorld~\cite{trivedi2024appworld}, and \underline{Task Score}
(TS) and \underline{Success Rate} (SR) on WebShop~\cite{yao2022webshop}.
Inference cost is measured by the total input and output tokens
consumed by the entire team, including auxiliary model calls,
and normalized to the Independent baseline.
Non-model storage and retrieval costs are excluded for all methods.
Figure~\ref{fig:token_costs} displays
$100(\bar C_m/\bar C_{\mathrm{Independent}}-1)\%$ from unrounded
token means; negative values indicate fewer tokens. Ratio uncertainty
is not estimated from the aggregate summaries.
% Exploration diversity is measured by the average pairwise
% Jaccard distance~\cite{kosub2019jaccard} between agents' accessed resources,
% using API endpoints for AppWorld and product IDs for WebShop.

\noindent\paragraph{\textbf{Implementation Details}.}
Unless otherwise specified, we use three parallel agents
solving the same task in separate environments.
We evaluate Qwen3.5-35B-A3B and Qwen3.5-4B~\cite{qwen2026qwen35} as task models.
All task and answering models remain frozen, while only
the Qwen3-0.6B gate~\cite{qwen2025qwen3} is trained using trajectories collected
with Qwen3.5-35B-A3B on both benchmarks.
The trained gate is shared across task models within each benchmark.

For final evaluation, each agent reports its confidence,
and we select the eligible candidate with the highest confidence
without accessing benchmark rewards.
We also report \underline{Oracle@3}, which identifies the best outcome
among the three candidates using benchmark feedback.
The former reflects performance under a practical selection
strategy, while the latter measures the potential of the
generated candidates and reveals limitations in final selection.
All experiments are repeated with three generation seeds.
Further details are provided in
the supplementary material.

\subsection{Main Results}
\label{sec:main_results}

\noindent\textbf{AOD consistently improves task performance.}
As shown in Table~\ref{tab:main_results}, \ours{}
outperforms Independent on both metrics across all
model and benchmark settings. The improvement is
particularly evident on AppWorld, where TGC increases
by 8.89 percentage points with Qwen3.5-35B-A3B.
Compared with other communication methods, \ours{}
achieves the best selected performance in three of
the four settings, with LTS (Adapted) leading on 4B AppWorld.
These results highlight the benefits of accessing
peer information when needed, beyond independent
parallel execution.

\noindent\textbf{AOD achieves a favorable balance between
performance and inference cost.}
Figure~\ref{fig:token_costs} shows that \ours{}
uses fewer tokens than PeriodicShare and LTS (Adapted)
across all four settings. It also improves task
performance over Independent while reducing token
consumption on 35B WebShop and 4B AppWorld.
Agentfugue (Adapted) uses fewer tokens than \ours{} in three
settings. It additionally replaces completed execution
segments in the agent context with summaries and retains
the raw segments in an external hub~\cite{hu2026agentfugue}.
This reduces repeated processing of full histories.
The other methods do not use this compression, so the
token comparison does not hold context management fixed
and cannot isolate communication efficiency.

\noindent\textbf{Oracle results reveal the importance
of final selection.}
On 4B AppWorld, \ours{} achieves the highest
Oracle@3 TGC, although LTS (Adapted) obtains better
results under confidence selection.
The larger gap between Oracle@3 and selected TGC
for \ours{} indicates that promising candidates
are not always recognized by the final selector.
This highlights the potential for improved selection
strategies to further enhance the effectiveness
of parallel agentic systems.

% BEGIN GENERATED: figures/token_costs.tex
% Generated by experiments/render_costs.py; edit main_results.csv.
\begin{figure}[t]
\centering
\includegraphics[width=\columnwidth]{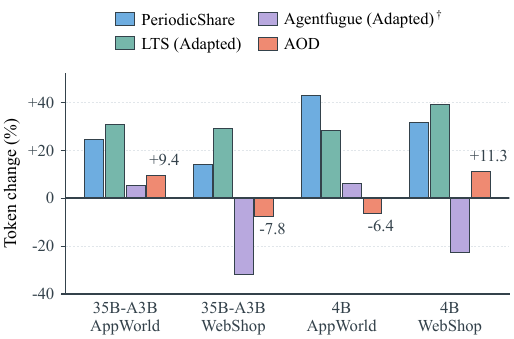}
\caption{\textbf{Full-team token change} relative to Independent
(three-seed means; lower is better).
$\dagger$Agentfugue (Adapted) additionally compresses completed context, storing raw
histories in an external hub; context management differs across methods.}
\Description{Four groups of vertical bars compare relative token changes for
PeriodicShare, LTS (Adapted), Agentfugue (Adapted), and AOD with two models on two benchmarks.
AOD is below PeriodicShare and LTS (Adapted) in all groups, and below
Independent on 35B WebShop and 4B AppWorld. Agentfugue (Adapted) is below AOD
in three groups and uses additional context compression and hub storage.}
\label{fig:token_costs}
\end{figure}
% END GENERATED: figures/token_costs.tex

% \noindent\textbf{Exploration Diversity.}
% Figure~\ref{fig:parallel_exploration} examines exploration
% diversity and resource coverage across parallel agents.
% On AppWorld, \ours{} achieves greater diversity than
% Independent while maintaining similar coverage.
% On WebShop, it exhibits lower diversity but broader
% coverage, indicating that agents collectively explore
% more distinct products despite greater overlap.
% These findings suggest that the performance benefits
% of on-demand communication do not require a uniform
% increase in exploration diversity.

\begin{figure*}[t]
    \centering
    \includegraphics[width=\textwidth]{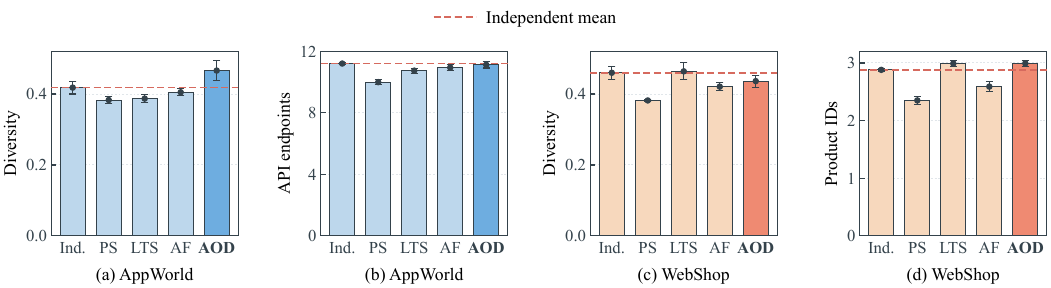}
    \caption{\textbf{Exploration diversity and coverage} with Qwen3.5-35B-A3B
    and three agents (mean $\pm$ standard deviation).
    Dashed lines mark Independent means. Ind., PS, LTS, and AF denote
    Independent, PeriodicShare, LTS (Adapted), and Agentfugue (Adapted).}
    \Description{Four vertical bar panels compare Independent,
    PeriodicShare, LTS (Adapted), Agentfugue (Adapted), and AOD. From left to right:
    AppWorld diversity, AppWorld API-endpoint coverage, WebShop diversity,
    and WebShop product-ID coverage. AppWorld bars are blue and WebShop
    bars are warm orange, with darker blue and coral bars for AOD. Dark points and error bars
    show means and standard deviations; coral dashed lines mark
    Independent means. All vertical axes start at zero.}
    \label{fig:parallel_exploration}
\end{figure*}

\subsection{Why Is AOD Effective?}
\label{sec:why_effective}

We investigate the effectiveness of \ours{} by analyzing
its communication decisions, peer selection and answering,
and exploration behavior, supplemented by ablation studies.

\noindent\textbf{Setup.}
We conduct our analysis using Qwen3.5-35B-A3B with
three parallel agents on AppWorld and WebShop.
To characterize parallel exploration, we consider the
resources accessed by each agent (API endpoints in AppWorld
and product IDs in WebShop).
Let $E_i$ denote the set of distinct resources accessed
by agent $i$. We introduce two complementary metrics:
1) \underline{Exploration Diversity} ($D$), which measures
the differences between agents' exploration trajectories
using the average pairwise Jaccard distance~\cite{kosub2019jaccard}:
\begin{equation}
D = \frac{2}{K(K-1)} \sum_{i<j}
\left(1-\frac{|E_i\cap E_j|}{|E_i\cup E_j|}\right),
\label{eq:exploration_diversity}
\end{equation}
where the distance is defined as zero for two empty sets;
and 2) \underline{Team Coverage} ($\mathrm{Cov}$), which measures
the total number of distinct resources explored by the
team, defined as $\mathrm{Cov} = |\bigcup_{i=1}^{K}E_i|$.

% BEGIN GENERATED: tables/gate_ablation.tex
% Generated by experiments/render_tables.py; edit the CSV, not this file.
\begin{table}[!t]
\centering
\caption{\textbf{Help-seeking gate ablation} with Qwen3.5-35B-A3B.}
\label{tab:gate_ablation}
\footnotesize
\renewcommand{\arraystretch}{1.18}
\setlength{\tabcolsep}{2pt}
\begin{tabularx}{\columnwidth}{l*{3}{>{\centering\arraybackslash}X}}
\toprule[0.7pt]
\rowcolor{PaperTableHeader}
\multicolumn{4}{c}{\textbf{AppWorld}} \\
\rowcolor{PaperTableHeader}
\textbf{Policy} & \textbf{TGC}$\uparrow$ & \textbf{SGC}$\uparrow$ & \textbf{Cost}$\downarrow$ \\
\midrule[0.35pt]
Autonomous & \expval{56.67} & \expval{31.67} & \expval{\ensuremath{\times1.13}} \\
AOD-Prompt & \expval{\underline{57.50}} & \expval{\underline{33.33}} & \expval{\ensuremath{\times\underline{1.11}}} \\
\cellcolor{PaperAODTint}\textbf{AOD} & \cellcolor{PaperAODTint}\expval{\textbf{62.78}} & \cellcolor{PaperAODTint}\expval{\textbf{38.33}} & \cellcolor{PaperAODTint}\expval{\ensuremath{\times\mathbf{1.09}}} \\
\midrule[0.35pt]
\rowcolor{PaperTableHeader}
\multicolumn{4}{c}{\textbf{WebShop}} \\
\rowcolor{PaperTableHeader}
\textbf{Policy} & \textbf{TS}$\uparrow$ & \textbf{SR}$\uparrow$ & \textbf{Cost}$\downarrow$ \\
\midrule[0.35pt]
Autonomous & \expval{68.60} & \expval{44.00} & \expval{\ensuremath{\times\underline{0.98}}} \\
AOD-Prompt & \expval{\underline{69.59}} & \expval{\underline{45.00}} & \expval{\ensuremath{\times1.00}} \\
\cellcolor{PaperAODTint}\textbf{AOD} & \cellcolor{PaperAODTint}\expval{\textbf{70.27}} & \cellcolor{PaperAODTint}\expval{\textbf{45.50}} & \cellcolor{PaperAODTint}\expval{\ensuremath{\times\mathbf{0.92}}} \\
\bottomrule[0.7pt]
\end{tabularx}
\end{table}
% END GENERATED: tables/gate_ablation.tex

% BEGIN GENERATED: tables/peer_selection.tex
% Generated by experiments/render_tables.py; edit peer_selection.csv.
\begin{table}[!t]
\centering
\caption{\textbf{Peer selection and answering ablation}
with Qwen3.5-35B-A3B.}
\label{tab:peer_selection}
\footnotesize
\renewcommand{\arraystretch}{1.18}
\setlength{\tabcolsep}{2pt}
\begin{tabularx}{\columnwidth}{l*{3}{>{\centering\arraybackslash}X}}
\toprule[0.7pt]
\rowcolor{PaperTableHeader}
\multicolumn{4}{c}{\textbf{AppWorld}} \\
\rowcolor{PaperTableHeader}
\textbf{Variant} & \textbf{TGC}$\uparrow$ & \textbf{SGC}$\uparrow$ & \textbf{Cost}$\downarrow$ \\
\midrule[0.35pt]
Random Peer & \expval{58.33} & \expval{30.00} & \expval{\ensuremath{\times\underline{1.13}}} \\
Evidence-Based & \expval{\underline{60.00}} & \expval{\textbf{40.50}} & \expval{\ensuremath{\times1.15}} \\
\cellcolor{PaperAODTint}\textbf{AOD} & \cellcolor{PaperAODTint}\expval{\textbf{62.78}} & \cellcolor{PaperAODTint}\expval{\underline{38.33}} & \cellcolor{PaperAODTint}\expval{\ensuremath{\times\mathbf{1.09}}} \\
\midrule[0.35pt]
\rowcolor{PaperTableHeader}
\multicolumn{4}{c}{\textbf{WebShop}} \\
\rowcolor{PaperTableHeader}
\textbf{Variant} & \textbf{TS}$\uparrow$ & \textbf{SR}$\uparrow$ & \textbf{Cost}$\downarrow$ \\
\midrule[0.35pt]
Random Peer & \expval{66.16} & \expval{40.50} & \expval{\ensuremath{\times\underline{0.97}}} \\
Evidence-Based & \expval{\underline{68.58}} & \expval{\underline{41.50}} & \expval{\ensuremath{\times0.98}} \\
\cellcolor{PaperAODTint}\textbf{AOD} & \cellcolor{PaperAODTint}\expval{\textbf{70.27}} & \cellcolor{PaperAODTint}\expval{\textbf{45.50}} & \cellcolor{PaperAODTint}\expval{\ensuremath{\times\mathbf{0.92}}} \\
\bottomrule[0.7pt]
\end{tabularx}
\end{table}
% END GENERATED: tables/peer_selection.tex

\subsubsection{Help-Seeking Gate}

We evaluate the learned gate by comparing three policies
in Table~\ref{tab:gate_ablation}:
1) \underline{Autonomous}, which executes all proposed requests;
2) \underline{AOD-Prompt}, which uses the pretrained
Qwen3-0.6B gate; and
3) \underline{AOD}, which learns approval decisions from
downstream team outcomes.
All variants share the same request formulation,
peer selection, and answering mechanisms.

\noindent\textbf{Selective help seeking improves performance
and efficiency.}
As shown in Table~\ref{tab:gate_ablation},
\ours{} outperforms both alternatives across both benchmarks
while consuming fewer tokens.
On AppWorld, TGC improves by 5.28 percentage points
over AOD-Prompt, alongside higher SGC.
On WebShop, performance improvements accompany a reduction
in relative token cost from $\times 1.00$ to $\times 0.92$.
% Notably, the trained gate executes approximately 62\%
% fewer peer requests on both benchmarks.
These results highlight the value of learning which
requests warrant communication, rather than simply
increasing the frequency of peer assistance.

% Declare the wide scaling table early so it stays near its discussion.
% BEGIN GENERATED: tables/scaling.tex
% Generated by experiments/render_scaling.py; edit scaling_results.csv.
\begin{table*}[t]
\centering
\caption{\textbf{Scaling} with Qwen3.5-35B-A3B.}
\label{tab:scaling}
\footnotesize
\renewcommand{\arraystretch}{1.18}
\setlength{\tabcolsep}{3pt}
\begin{tabularx}{\textwidth}{cl*{10}{>{\centering\arraybackslash}X}}
\toprule[0.7pt]
\rowcolor{PaperTableHeader}
&
& \multicolumn{5}{c}{\textbf{AppWorld}} & \multicolumn{5}{c}{\textbf{WebShop}} \\
\rowcolor{PaperTableHeader}
& & &
& \multicolumn{2}{c}{\textbf{Oracle@}$K$}
& & &
& \multicolumn{2}{c}{\textbf{Oracle@}$K$}
& \\
\rowcolor{PaperTableHeader}
\multirow{-3}{*}{\textbf{$K$}} & \multirow{-3}{*}{\textbf{Method}}
& \multirow{-2}{*}{\textbf{TGC}$\uparrow$}
& \multirow{-2}{*}{\textbf{SGC}$\uparrow$}
& \textbf{TGC}$\uparrow$ & \textbf{SGC}$\uparrow$
& \multirow{-2}{*}{\shortstack{\textbf{Tokens}\\\textbf{(M/task)}$\downarrow$}}
& \multirow{-2}{*}{\textbf{TS}$\uparrow$}
& \multirow{-2}{*}{\textbf{SR}$\uparrow$}
& \textbf{TS}$\uparrow$ & \textbf{SR}$\uparrow$
& \multirow{-2}{*}{\shortstack{\textbf{Tokens}\\\textbf{(M/task)}$\downarrow$}} \\
\midrule[0.35pt]
\multirow{2}{*}{5} & Independent & \expval{49.17} & \expval{30.00} & \expval{69.17} & \expval{57.50} & \expval{\textbf{1.737}} & \expval{69.12} & \expval{43.00} & \expval{74.13} & \expval{49.50} & \expval{\textbf{0.695}} \\
 & \cellcolor{PaperAODTint}\textbf{AOD} & \cellcolor{PaperAODTint}\expval{\textbf{57.50}} & \cellcolor{PaperAODTint}\expval{\textbf{32.50}} & \cellcolor{PaperAODTint}\expval{\textbf{79.17}} & \cellcolor{PaperAODTint}\expval{\textbf{65.00}} & \cellcolor{PaperAODTint}\expval{1.883} & \cellcolor{PaperAODTint}\expval{\textbf{69.95}} & \cellcolor{PaperAODTint}\expval{\textbf{44.00}} & \cellcolor{PaperAODTint}\expval{\textbf{77.89}} & \cellcolor{PaperAODTint}\expval{\textbf{56.00}} & \cellcolor{PaperAODTint}\expval{0.730} \\
\midrule[0.35pt]
\multirow{2}{*}{7} & Independent & \expval{52.50} & \expval{25.00} & \expval{80.83} & \expval{65.00} & \expval{\textbf{2.448}} & \expval{70.03} & \expval{46.00} & \expval{78.41} & \expval{\textbf{56.00}} & \expval{0.816} \\
 & \cellcolor{PaperAODTint}\textbf{AOD} & \cellcolor{PaperAODTint}\expval{\textbf{64.17}} & \cellcolor{PaperAODTint}\expval{\textbf{42.50}} & \cellcolor{PaperAODTint}\expval{\textbf{82.50}} & \cellcolor{PaperAODTint}\expval{\textbf{70.00}} & \cellcolor{PaperAODTint}\expval{2.632} & \cellcolor{PaperAODTint}\expval{\textbf{71.67}} & \cellcolor{PaperAODTint}\expval{\textbf{48.00}} & \cellcolor{PaperAODTint}\expval{\textbf{78.49}} & \cellcolor{PaperAODTint}\expval{\textbf{56.00}} & \cellcolor{PaperAODTint}\expval{\textbf{0.748}} \\
\midrule[0.35pt]
\multirow{2}{*}{9} & Independent & \expval{55.83} & \expval{30.00} & \expval{88.33} & \expval{\textbf{82.50}} & \expval{\textbf{3.085}} & \expval{70.51} & \expval{\textbf{46.00}} & \expval{78.90} & \expval{56.50} & \expval{1.084} \\
 & \cellcolor{PaperAODTint}\textbf{AOD} & \cellcolor{PaperAODTint}\expval{\textbf{62.50}} & \cellcolor{PaperAODTint}\expval{\textbf{42.50}} & \cellcolor{PaperAODTint}\expval{\textbf{89.17}} & \cellcolor{PaperAODTint}\expval{\textbf{82.50}} & \cellcolor{PaperAODTint}\expval{3.247} & \cellcolor{PaperAODTint}\expval{\textbf{72.30}} & \cellcolor{PaperAODTint}\expval{44.50} & \cellcolor{PaperAODTint}\expval{\textbf{81.49}} & \cellcolor{PaperAODTint}\expval{\textbf{60.00}} & \cellcolor{PaperAODTint}\expval{\textbf{1.005}} \\
\bottomrule[0.7pt]
\end{tabularx}
\end{table*}
% END GENERATED: tables/scaling.tex

\subsubsection{Peer Selection and Answering}

\noindent\textbf{Setup.}
Table~\ref{tab:peer_selection} compares three variants:
1) \underline{Random Peer}, which randomly selects a peer to generate a response; 2) \underline{Evidence-Based}, which directly returns retrieved BM25 passages~\cite{robertson2009probabilistic}; and 3) \underline{\ours{}}, which selects a relevant peer using BM25 and generates a tailored response. All variants use the same trained gate.

\noindent\textbf{Relevant peer selection improves performance
and efficiency.}
As shown in Table~\ref{tab:peer_selection},
\ours{} outperforms Random Peer on both benchmarks,
with gains of 8.33 percentage points in AppWorld SGC
and 5.00 points in WebShop SR, while using fewer tokens.
With the gate and answering model fixed, these results
highlight the importance of consulting relevant peers.

\noindent\textbf{Targeted answering enhances information utilization.}
Compared with direct BM25 retrieval, \ours{} improves
both WebShop metrics and AppWorld TGC, although direct
retrieval achieves higher AppWorld SGC (40.50 versus 38.33).
Moreover, \ours{} consumes fewer total tokens on both
benchmarks despite the additional answering step.
Human evaluation on fixed requests also finds more relevant new
evidence and observed evidence use with AOD
(see the supplementary material).

\subsubsection{Exploration Diversity and Coverage}

\noindent\textbf{AOD balances information sharing and independent exploration.}
Figure~\ref{fig:parallel_exploration} reveals two distinct patterns across benchmarks. 1) \underline{Complementary exploration on AppWorld.} \ours{} achieves the highest mean diversity while maintaining coverage comparable to Independent. This suggests that targeted assistance allows agents to reuse peer discoveries without converging on identical exploration paths. 2) \underline{Broader collective search on WebShop.} Although \ours{} exhibits slightly lower diversity than Independent and LTS (Adapted), it covers more distinct products than Independent, with coverage comparable to LTS (Adapted). This indicates that greater overlap between agents can coexist with broader collective exploration.

Together, these findings highlight the importance of considering diversity and coverage jointly.
Higher diversity does not necessarily imply broader exploration. More importantly, \ours{} improves task performance under both patterns, suggesting that effective collaboration depends on balancing information reuse and independent exploration rather than maximizing diversity alone.

Observed communication and execution traces are
provided in the supplementary material.

\subsection{Sensitivity Analysis}
\label{sec:sensitivity}

We examine the effects of gate supervision and team size
using Qwen3.5-35B-A3B.

\subsubsection{Rollout Probability and Reward Target}
\label{sec:supervision_sensitivity}

\noindent\textbf{Setup.}
We examine two factors in gate supervision:
1) \underline{Rollout probability} ($p$), which controls
approval of subsequent requests during paired rollouts,
with $p\in\{0.25,0.5,0.75\}$ and the initial decision fixed;
and 2) \underline{Reward target}, comparing team reward
$R$ with requester reward $R_i=\mathcal{R}(\tau_i)$
at $p=0.5$.
Remaining team inference cost breaks reward ties.
Each configuration trains a separate gate, evaluated
with three agents and three seeds.

\noindent\textbf{Rollout probability affects gate performance.}
Figures~\ref{fig:sensitivity_appworld}
and~\ref{fig:sensitivity_webshop} show that $p=0.5$
offers the best overall balance.
On AppWorld, it achieves the highest TGC and SGC
with competitive inference cost.
On WebShop, $p=0.25$ yields a slightly higher Task Score,
while $p=0.5$ achieves the highest SR (45.50)
and lowest token consumption.
These results highlight the importance of accounting
for future assistance when learning communication decisions.

\noindent\textbf{Team reward better aligns supervision
with collective outcomes.}
Compared with requester reward, team reward improves
both performance metrics on both benchmarks,
with similar or lower inference cost.
Notably, WebShop SR increases from 42.50 to 45.50.
This suggests that effective communication should be
evaluated by its contribution to team outcomes,
rather than individual agent progress alone.

\subsubsection{Scaling to Larger Teams}
\label{sec:scaling}

\noindent\textbf{Setup.}
We compare \ours{} with Independent at team sizes
$K\in\{5,7,9\}$, reusing the trained gate without retraining.
Each comparison uses the same tasks, seeds, per-agent budgets,
and final selection rule.

\noindent\textbf{AOD scales effectively to nine agents.}
Across all tested team sizes, \ours{} improves AppWorld TGC
and SGC and WebShop Task Score over Independent
(Table~\ref{tab:scaling}). The AppWorld SGC gain reaches
17.50 percentage points at $K=7$. On WebShop, \ours{}
achieves higher Task Score with approximately 7--8\% fewer
total tokens at $K=7,9$. It also matches or exceeds Independent
on every Oracle metric across both benchmarks.
These gains show that the learned communication policy
transfers to larger teams without retraining.

\begin{figure}[t]
    \centering
    \includegraphics[width=\columnwidth]{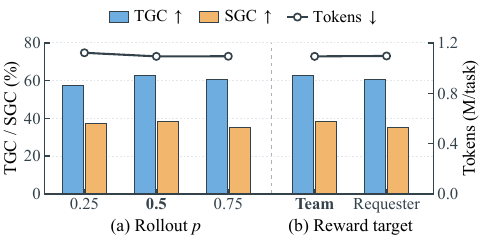}
    \caption{\textbf{Gate-training sensitivity on AppWorld.}
    Bars show mean performance over three seeds; markers show mean
    full-team tokens. Bold labels mark default settings.}
    \Description{Two adjacent groups compare AppWorld performance
    and token consumption: rollout probability on the left and
    reward target on the right, with shared axes and legend.
    The default probability of 0.5 achieves the highest TGC
    and SGC. Team reward improves performance over requester
    reward with similar token consumption.}
    \label{fig:sensitivity_appworld}
\end{figure}

\begin{figure}[t]
    \centering
    \includegraphics[width=\columnwidth]{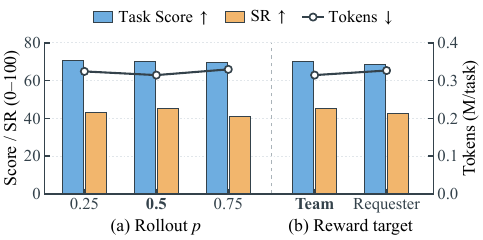}
    \caption{\textbf{Gate-training sensitivity on WebShop.}
    Bars show mean performance over three seeds; markers show mean
    full-team tokens. Bold labels mark default settings.}
    \Description{Two adjacent groups compare WebShop performance
    and token consumption: rollout probability on the left and
    reward target on the right, with shared axes and legend.
    The probability of 0.5 achieves the highest SR and lowest
    token consumption, while 0.25 achieves the highest Task
    Score. Team reward improves both metrics over requester
    reward while using fewer tokens.}
    \label{fig:sensitivity_webshop}
\end{figure}
\endgroup

\section{Conclusion}
\label{sec:conclusion}

We presented Ask on Demand (AOD), a communication framework
that enables parallel agents to seek assistance from peers
when specific information needs arise. We further train a
lightweight gate to identify requests worth executing,
using paired team continuations and supervision based on
final team outcomes and inference cost. Experiments on
AppWorld and WebShop demonstrate consistent performance
improvements over independent parallel execution, with
lower token consumption than periodic sharing and adapted
shared memory methods. Further analyses highlight the
importance of selective communication and relevant peer
selection, while showing that performance gains do not
require a uniform increase in exploration diversity.
These findings suggest that aligning communication with
agents' evolving information needs offers an effective
approach to improving collaboration in parallel agentic
systems.

{
    \small
    \bibliographystyle{ieeenat_fullname}
    \bibliography{introduction}
}

\end{document}